\pdfoutput=1

\documentclass[11pt]{article}

\PassOptionsToPackage{table,xcdraw}{xcolor}
\usepackage{EMNLP2023}
\usepackage{times}
\usepackage{latexsym}

\usepackage[T1]{fontenc}

\usepackage[utf8]{inputenc}

\usepackage{microtype}

\usepackage{inconsolata}
\usepackage{amsmath} 
\usepackage{amssymb} 
\usepackage{amsfonts} 
\usepackage{geometry} 
\usepackage{multicol} 
\usepackage{multirow}
\usepackage{enumitem} 
\usepackage{ragged2e} 
\usepackage{lipsum} 
\usepackage{amsmath}
\usepackage{soul}
\usepackage{mathtools}
\usepackage{booktabs}
\usepackage[table,xcdraw]{xcolor}
\usepackage[normalem]{ulem}
\definecolor{mygray}{RGB}{239,239,239} 
\sethlcolor{mygray}

\usepackage{tabularray}  
\usepackage{caption}     
\usepackage{siunitx}    
\usepackage{graphicx}
\usepackage{array}

\usepackage{amssymb}
\usepackage{url}

\usepackage{times}
\usepackage{latexsym}
\usepackage{tikz}

\title{ReasonGRM: Enhancing Generative Reward Models through Large Reasoning Models}

\author{Bin Chen\textsuperscript{1}\quad
      Xinzge Gao\textsuperscript{1}\quad
      Chuanrui Hu\textsuperscript{\dag 1}\thanks{\hspace{2mm}Project Leader.}\quad
      Penghang Yu\textsuperscript{1}\quad
      Hua Zhang\textsuperscript{1}\quad
      Bing-Kun Bao\textsuperscript{\ddag 2}\thanks{\hspace{2mm}Corresponding author.} \\
      \textsuperscript{1}Qihoo360 \\
      \textsuperscript{2}Nanjing University of Posts and Telecommunications, Nanjing, China \\
      \texttt{\{chenbin, gaoxingze, huchuanrui, yupenghang,  zhanghua\}@360.cn } \\
      \texttt{bingkunbao@njupt.edu.cn}
    }

\begin{document}
\maketitle

\begin{abstract}
Generative Reward Models (GRMs) provide greater flexibility than scalar reward models in capturing human preferences, but their effectiveness is limited by poor reasoning capabilities.
This often results in incomplete or overly speculative reasoning paths, leading to hallucinations or missing key information in complex tasks. We address this challenge with ReasonGRM, a three-stage generative reward modeling framework.
In the first stage, Zero-RL is used to generate concise, outcome-directed reasoning paths that reduce the likelihood of critical omissions.
In the second stage, we introduce a novel evaluation metric, $R^\star$, which scores reasoning paths based on their generation likelihood.
This favors paths that reach correct answers with minimal exploration, helping to reduce hallucination-prone data during training.
In the final stage, the model is further refined through reinforcement learning on challenging examples to enhance its preference discrimination capabilities.
Experiments on three public benchmarks show that ReasonGRM achieves competitive or state-of-the-art performance, outperforming previous best GRMs by 1.8\% on average and surpassing proprietary models such as GPT-4o by up to 5.6\%.
These results demonstrate the effectiveness of reasoning-aware training and highlight the importance of high-quality rationale selection for reliable preference modeling.

\end{abstract}

\section{Introduction}

\begin{figure}[t!]
  \centering 
  \includegraphics[width=1.0\linewidth]{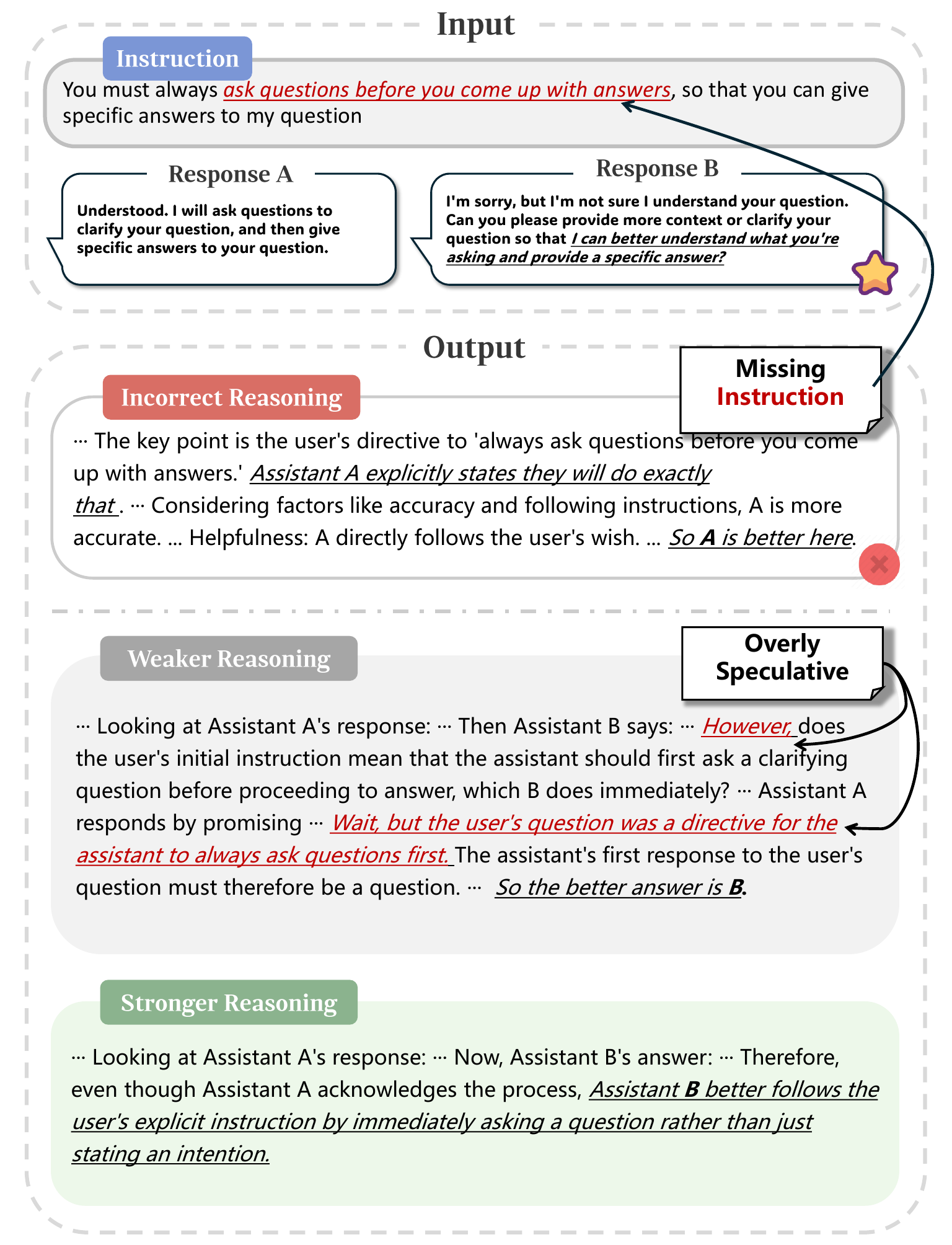}
  \caption{The figure delineates three reasoning pathways in data generation: 
    \textbf{1)Incorrect Reasoning} wrongly favors an assistant who says they will ask questions over one who actually asks one immediately. \textbf{2) Weaker Reasoning} reach the correct answer but through a convoluted or uncertain process, making it "suboptimal" for learning, and \textbf{3) Stronger Reasoning} Clearly and directly identifies the better response by focusing on the assistant's immediate adherence to the explicit instruction. This is  better for learning. }
  \label{fig:sample} 
\end{figure}


LLMs, such as GPT~\cite{achiam2023gpt}, Claude~\cite{claude3}, DeepSeek~\cite{deepseekai2025deepseekv3technicalreport, deepseekai2025deepseekr1incentivizingreasoningcapability}, Qwen~\cite{yang2024qwen2technicalreport, qwen2025qwen25technicalreport}, and Llama~\cite{grattafiori2024llama3herdmodels}, have achieved impressive advancements in understanding, generation, and decision-making, significantly enhancing the generalization capabilities of AI systems.
However, reliably deploying these models in real-world applications hinges on aligning their outputs with human values.
Reinforcement Learning from Human Feedback (RLHF)~\cite{ouyang2022traininglanguagemodelsfollow, dong2024rlhfworkflowrewardmodeling} has emerged as the dominant framework to address this challenge, with the Reward Model (RM) playing a central role in aligning model behavior with human preferences.

Conventional Scalar Reward Models (SRMs) \cite{wang2024helpsteer2, liu2024skywork} operate under a fixed input paradigm and often overfit to training data, making them inflexible in adapting to different domains or personalized preference profiles.
In contrast, Generative Reward Models (GRMs)\cite{zheng2023judgingllmasajudgemtbenchchatbot} exploit prompt design to leverage the rich reasoning and decision-making capabilities of LLMs, enabling more expressive and adaptable preference modeling.
As a result, GRMs are increasingly becoming the preferred choice for both research and deployment~\cite{kimiteam2025kimik15scalingreinforcement, openai2024openaio1card}.

Recent research suggests that incorporating reasoning steps can enhance reward model performance~\cite{ankner2024critiqueoutloudrewardmodels, wang2024selftaughtevaluators, mahan2024generativerewardmodels}.
However, they often treat reasoning as inherently beneficial without explicitly defining what constitutes desirable reasoning for preference alignment.
Consequently, the generated or collected rationales are frequently noisy, logically inconsistent, or misaligned with task objectives—undermining both the effectiveness and stability of GRM training.
This gap motivates a fundamental question:
\textbf{What type of reasoning is truly necessary for preference modeling?}

To address this, we examine the characteristics of reasoning that yield robust and reliable preference judgments.
We propose that high-quality reasoning paths must satisfy two key properties.
First, they must lead to the correct outcome—regardless of length or format—a criterion we term Validity.
Second, they must exhibit internal coherence and logical continuity, avoiding speculative or erratic detours, which we define as Self-Consistency.
Together, these criteria ensure that reasoning is both outcome-aligned and structurally sound.

Based on this, we introduce $R^\star$, a simple yet effective metric that jointly evaluates Validity and Self-Consistency through generation likelihood.
By scoring rationales based on their likelihood under a preference-aware LLM, $R^\star$ enables the automatic selection of high-quality reasoning paths from noisy candidate sets.
This selection process forms the foundation of ReasonGRM, our proposed generative reward modeling framework.

ReasonGRM consists of three stages:
(1) generating diverse reasoning paths using Zero-RL;
(2) filtering these paths with $R^\star$ to construct high-quality training data; and
(3) fine-tuning the model using both supervised learning and reinforcement learning to enhance its preference discrimination capabilities.

We conduct extensive experiments on three widely used benchmarks: RM-Bench~\cite{liu2024rm}, RewardBench~\cite{lambert2024rewardbenchevaluatingrewardmodels}, and RMB~\cite{zhou2024rmb}. The results show that ReasonGRM achieves state-of-the-art performance, outperforming the strongest open-source GRM by 1.8\%, GPT-4o by 5.6\%, and the leading SRM by 4.5\% on average. These results validate the effectiveness of our approach. Furthermore, we conduct detailed ablation studies to analyze the role of reasoning quality in model performance, the effectiveness and generalizability of $R^\star$ for rationale filtering, and the impact of each training stage on the final reward model.
In conclusion, our main contributions are as follows:
\begin{itemize}
    \item We demonstrate the critical role of reasoning quality in generative reward models and propose ReasonGRM, a reasoning-aware framework that significantly improves preference modeling performance.
    \item We introduce $R^\star$, a novel metric for evaluating and filtering high-quality reasoning paths, effectively addressing the data quality bottleneck in training reward models for complex tasks.
    \item We design a multi-stage training pipeline that transforms a general-purpose LLM into a high-performing, preference-specialized reward model.
    \item We conduct comprehensive experiments across multiple benchmarks, where ReasonGRM achieves state-of-the-art performance, validating both its effectiveness and superiority over existing generative and scalar reward models.
\end{itemize}

\section{Related Works}
\subsection{Reward Models}
Early reward models were predominantly SRMs, which are typically fine-tuned from pre-trained language models by replacing their final layer with a regression head~\cite{liu2024skywork}. SRMs have demonstrated efficacy in numerous alignment tasks and have been utilized in RLHF  for multiple LLMs. However, the compression of complex, multidimensional human preferences into a singular scalar value by SRMs inevitably results in information loss, thereby hindering the capture of subtle nuances and specific rationales underlying these preferences. Consequently, SRMs exhibit limited flexibility and generalization capabilities when the evaluation focus requires dynamic adjustment across varying tasks~\cite{liu2025inferencetimescalinggeneralistreward}.

To address these limitations, researchers have proposed GRMs~\cite{mahan2024generativerewardmodels}. GRMs directly leverage the inherent generative capabilities of LLMs to articulate preferences. By inheriting the robust semantic understanding and contextual awareness of LLMs, GRMs also provide enhanced interpretability and flexibility, allowing them to cater to a broader range of evaluation demands~\cite{zhang2025generativeverifiersrewardmodeling}.Despite the considerable potential demonstrated by GRMs, their performance continues to face challenges, particularly in tasks that necessitate profound understanding and intricate judgment. A critical factor contributing to this is the inherent limitation in the reasoning abilities of the models themselves. This observation has consequently directed our attention towards the problem of reasoning within reward models.

\subsection{Reasoning in Reward Models}
The introduction of explicit reasoning steps into general LLMs has been demonstrated to significantly enhance their performance on complex tasks such as arithmetic, commonsense, and symbolic reasoning~\cite{wei2023chainofthoughtpromptingelicitsreasoning}. Inspired by these advancements, researchers have begun to explore the integration of reasoning mechanisms into RMs. For instance, ~\cite{liu2025inferencetimescalinggeneralistreward} improves the accuracy of reward scoring by guiding the model to generate structured "principles" and "critiques," which are then combined with an inference-time expansion strategy. ~\cite{liu2025inferencetimescalinggeneralistreward} designed outcome-driven rewards for discriminative tasks and utilized RL to endow the RM with reasoning capabilities, thereby enhancing its preference proficiency. Furthermore, other research, such as ~\cite{chen2025rmr1rewardmodelingreasoning}, has attempted to optimize reward models by distilling high-quality reasoning data from powerful proprietary teacher models.

In contrast, these existing works tend to focus on optimizing specific output paradigms, rely on feedback from final task performance, or leverage external teacher models. Our research, however, concentrates more directly on enhancing the quality of the General GRM's internal reasoning process. We aim to enable the model to reason independently within preference tasks by optimizing the data generation pipeline. This approach is intended to fundamentally strengthen the GRM's reasoning capabilities and the reliability of its judgment criteria, without dependence on guidance from external proprietary models.

\begin{figure*}[t]
\vspace{-0.3cm}
\setlength{\belowcaptionskip}{0cm}
\setlength{\abovecaptionskip}{0.1cm}
    \centering
    \includegraphics[width=0.9\linewidth]{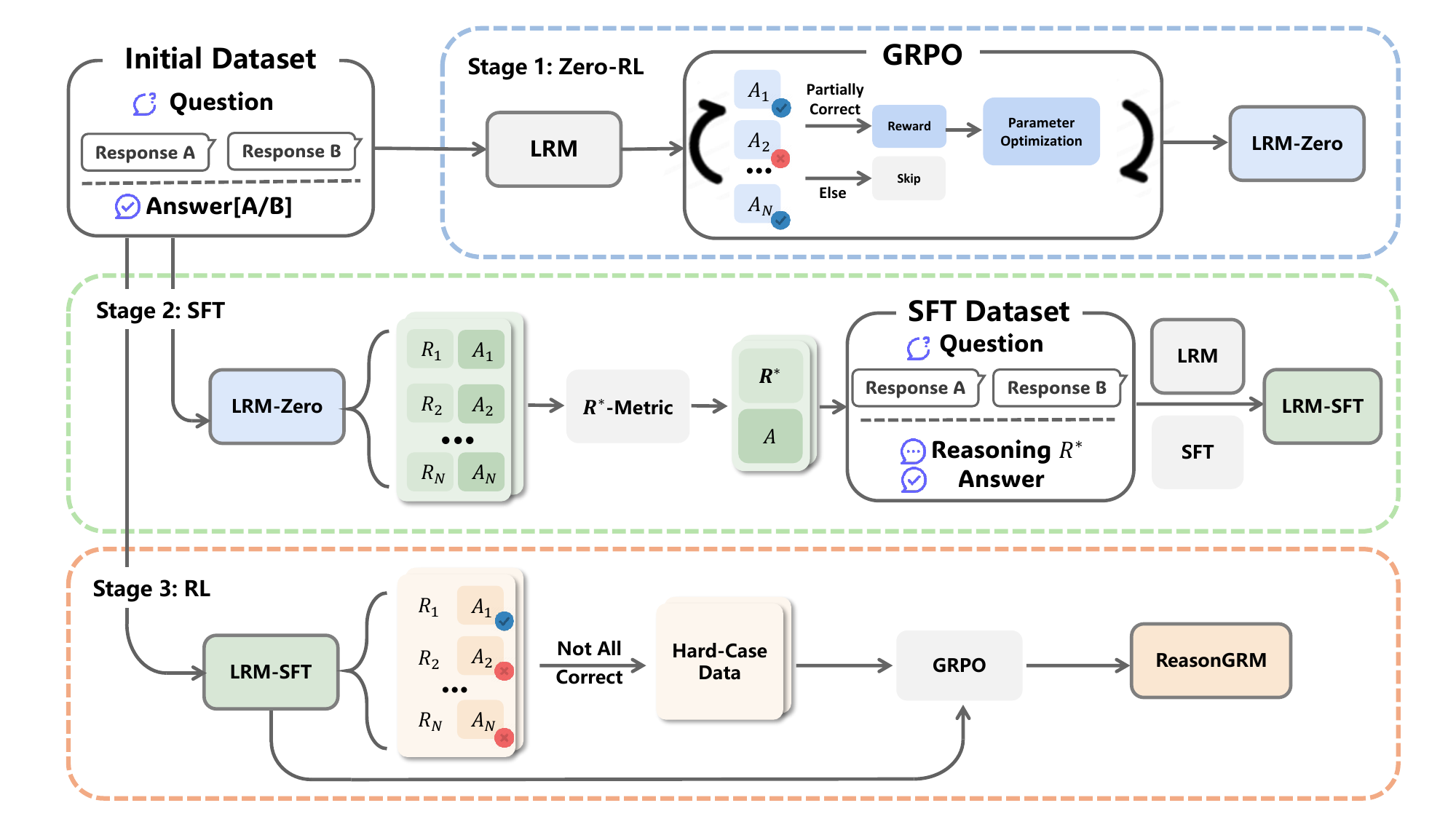}
    \caption{\textbf{Overview of the ReasonGRM training pipeline.}The process begins with an LRM adapted via GRPO for preference (LRM-Zero). Subsequently, LRM-SFT is developed using SFT with $^\star$-filtered reasoning from LRM-Zero, and ReasonGRM is finalized through GRPO-based RL refinement of LRM-SFT on hard cases.}
    \label{fig:overall}
    \vspace{-0.2cm}
\end{figure*}

\section{ReasonGRM}
We have previously discussed the importance of reasoning data for GRMs. However, when a foundation model possesses insufficient intrinsic reasoning capabilities, it becomes challenging to consistently generate accurate and high-quality reasoning data, even with extensive sampling. Furthermore, even when multiple reasoning paths leading to the correct answer are acquired, these paths may still exhibit significant disparities in their logical soundness, conciseness, and depth. Current methodologies frequently lack robust mechanisms for the meticulous evaluation and selection of the optimal reasoning path. Accordingly, this chapter focuses on two key aspects: 1) the generation of high-quality reasoning path data, and 2) the evaluation of the quality of these generated reasoning paths.

\subsection{Overall Architecture}
Inspired by the cold start of DeepSeek-R1~\cite{deepseekai2025deepseekr1incentivizingreasoningcapability}, our overall framework can be broadly divided into the following three stages as shown in Figure \ref{fig:overall}:
\begin{itemize}
    \item First, we utilize existing LRMs to train a Zero model, initially adapting it to the preference task. This Zero model is subsequently used to generate high-quality inference paths for the RM task.
    \item Subsequently, we design an inference path evaluation metric, $R^\star$, to identify and select high-quality inference paths from those generated by the Zero model. These selected paths are then used for supervised fine-tuning.
    \item Finally, we further fine-tune the model via reinforcement learning, specifically focusing on hard samples, to maximize its performance on the preference task.
\end{itemize}

\subsection{Zero-RL : Generate better $R$}
To enable the model to generate high-quality reasoning paths, it must first possess a fundamental understanding of the preference task's ultimate goal: to accurately identify the better answer. However, the initial dataset used for this stage only contains the question, candidate answers, and the correct final choice , without providing any explicit reasoning chains. The core objective of this stage is to perform an initial preference alignment of the base Large Reasoning Model (LRM) through reinforcement learning. This process teaches the model to make correct choices based solely on the final outcome, without being shown any demonstrative reasoning. The resulting model, now adept at recognizing correct outcomes, is termed LRM-Zero.

To achieve this, we employ the Generalized Reward Policy Optimization (GRPO) algorithm as follows:
\begin{itemize}
    \item \textbf{Policy Model}: The base LRM $M_{\pi_0}$, serves as the policy model to be optimized.
    \item \textbf{Reward}: Our reward mechanism is \textbf{outcome-driven}. A positive reward is granted if the model's generated answer matches the ground-truth answer $a_Q$ from the dataset; otherwise, no reward is given.
\end{itemize}

Furthermore, to enhance training efficiency and stability, we introduce a sample filtering mechanism. During training, we prompt the model to generate $K$ responses for a single question. If all $K$ responses are "entirely correct" (indicating the sample is too easy and provides a weak learning signal) or "entirely incorrect" (indicating the sample is too difficult, where forced learning could destabilize the model), we skip the parameter update for that sample in the current iteration. The GRPO update is performed if and only if the following condition is met:
$$
0 < \sum_{i=1}^{K} \mathbb{I}(\text{is\_equivalent}(o_i, a_Q)) < K
$$
Where:
\begin{itemize}
    \item $\mathbb{I}(\cdot)$ is the indicator function, which is 1 if the condition is true, and 0 otherwise.
    \item $o_i$ is the $i$-th answer generated by the model.
    \item $a_Q$ is the ground-truth for the question.
    \item $K$ is the total number of generated responses.
\end{itemize}

This selective update strategy ensures that training efforts are concentrated on the most informative samples. These instances, where the model's output is inconsistent, provide the most effective learning gradients for refining its discriminative capabilities. By filtering out uninformative samples that are either trivial or excessively difficult, this approach leads to a more efficient and stable optimization process.

Upon completing this stage, the resulting \textbf{LRM-Zero}, while not having been trained on any explicit reasoning text, has acquired a strong initial capability for preference discrimination. This judge provides a solid foundation for generating high-quality and logically coherent reasoning paths.

\begin{figure*}[t]
\vspace{-0.3cm}
\setlength{\belowcaptionskip}{0cm}
\setlength{\abovecaptionskip}{0.1cm}
    \centering
    \includegraphics[width=\linewidth]{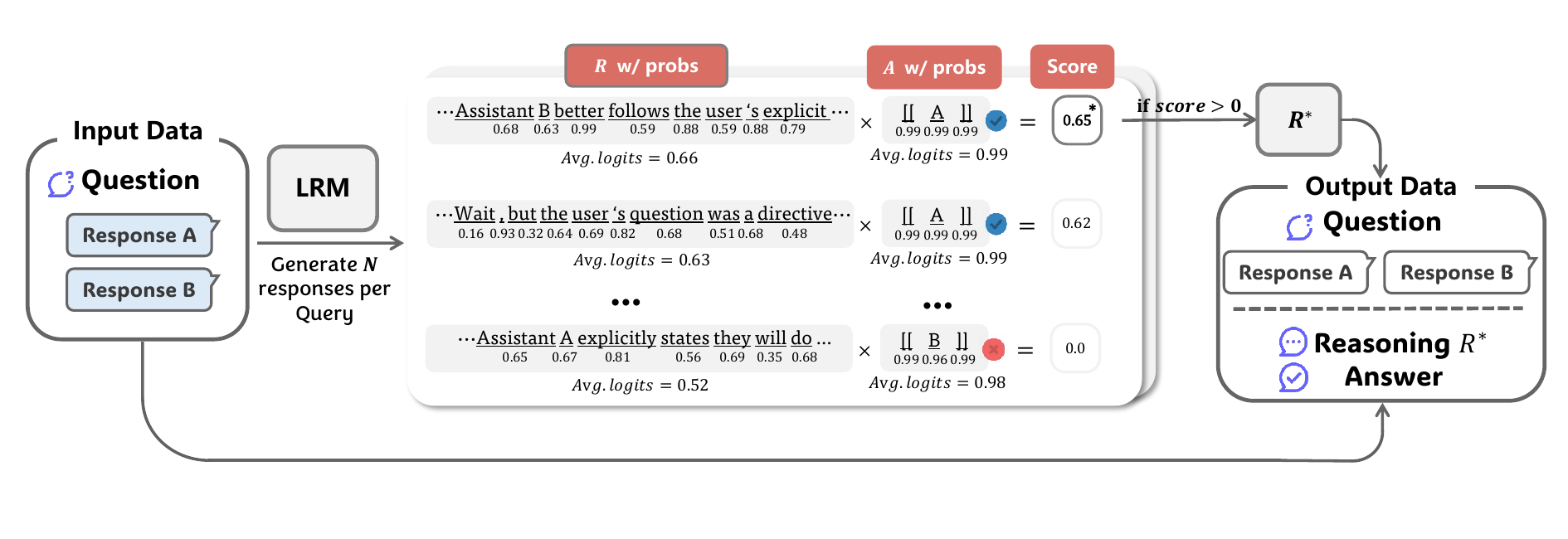}
    \caption{\textbf{Workflow of extracting stronger $R$ base on $R^\star$-Metric.}}
    \label{fig:Rstar}
    \vspace{-0.2cm}
\end{figure*}

\subsection{$R^\star$ : Extract stronger $R$}
High-quality training data is crucial for enhancing a model's reasoning capabilities. However, even when models can generate reasoning paths that yield the correct final answer, these paths may still vary significantly in logical clarity, conciseness, and confidence. To select the optimal paths from these preliminarily correct reasoning processes, we propose a metric for evaluating the quality of reasoning paths, termed $R^\star$. The core principle of $R^\star$ is that a superior reasoning path $R'$ should not only efficiently guide the model to produce the correct answer $A$, but 
the path $R'$ itself should also be confident and demonstrate a high degree of internal logical consistency.

The complete workflow for filtering high-quality reasoning paths using the $R^\star$ metric is illustrated in Figure \ref{fig:Rstar}. It begins with candidate generation by LRM-Zero model $M_Z$, roceeds through filtering via the $R^\star$ scoring mechanism, and culminates in the construction of a dataset for SFT.

First, for each question $Q^{(j)}$ within the input dataset, the $M_Z$ model is utilized to generate $G$ `reasoning-answer' pairs: $\{(R'_1, a_1), (R'_2, a_2), \dots ,(R'_G, a_G)\}$. Each reasoning path $R'_k$ is composed of a token sequence $(t'_{k,1}, \dots, t'_{k,L_{R'_k}})$, with its corresponding conditional token generation probability sequence being $\mathcal{P}_{R'_k} = (p'_{k,1}, \dots, p'_{k,L_{R'_k}})$; similarly, each answer $a_k$ comprises a token sequence $(t''_{k,1}, \dots, t''_{k,L_{a_k}})$, with its associated conditional token generation probability sequence $\mathcal{P}_{a_k} = (p''_{k,1}, \dots, p''_{k,L_{a_k}})$. We retain only those instances where the candidate answer $a_k$ aligns with the ground-truth answer $A^{(j)}$, thus forming an evaluation set $\mathcal{G}^{(j)}$ for each question. For any element $g$ within this set, its associated reasoning path is denoted by $R'_g$ (with token probabilities $p'_{g,i}$), and its answer by $a_g$ (with token probabilities $p''_{g,i}$).

Second, for every valid reasoning path $R'_{g}$ in $\mathcal{G}^{(j)}$ and its corresponding correct answer $a_g$, we calculate its $R^{*}$ score. This score is obtained by multiplying the `Self-Consistency' and `Validity' of the reasoning path:
$$
\begin{aligned}
    R^{*}(R'_{g}, a_g, Q) &= \text{P}(R'_{g} | Q) \cdot \text{P}(a_g | Q, R'_{g})  \\
                          &= \underbrace{\frac{1}{L_{R'_g}} \sum_{i=1}^{L_{R'_g}} p'_{g,i}}_{\text{Self-Consistency}} \cdot \underbrace{\frac{1}{L_{a_g}} \sum_{i=1}^{L_{a_g}} p''_{g,i}}_{\text{Validity}}
\end{aligned}
$$

Finally, for each question $Q^{(j)}$, the reasoning path $R'^{*(j)}$ that achieves the highest $R^{*}$ score is selected from $\mathcal{G}^{(j)}$. These selected optimal `reasoning-answer' pairs, $\{(Q^{(j)}, R'^{*(j)}, A^{(j)})\}_{j=1}^{K'}$, collectively form the high-quality SFT dataset $\mathcal{D}_{\text{SFT}}$. This resultant dataset, $\mathcal{D}_{\text{SFT}}$, is subsequently employed for the supervised fine-tuning of the base LRM $M_{\pi_0}$. This process yields the $M_{\text{SFT}}$ model, characterized by significantly enhanced reasoning capabilities.

\begin{table*}[h]
\vspace{0.4cm}
    \setlength{\abovecaptionskip}{0.15cm}
    \setlength{\belowcaptionskip}{0cm}
    \centering
    \resizebox{140mm}{!}{\begin{tabular}{@{}lcccc@{}}
\toprule
\multicolumn{1}{l|}{\textbf{Models}}                                            & \textbf{RewardBench} & \textbf{RM-Bench} & \multicolumn{1}{c|}{\textbf{RMB}}  & \textbf{Average} \\ \midrule
\multicolumn{5}{l}{\textit{\textbf{Scalar Reward Models}}}                                                                                                                         \\ \midrule
\multicolumn{1}{l|}{Skywork-Reward-Gemma-2-27B}                                 & {\underline{93.8}}           & 67.3              & \multicolumn{1}{c|}{60.2}          & 73.8             \\
\multicolumn{1}{l|}{Internlm2-7b-reward}                                        & 87.6                 & 67.1              & \multicolumn{1}{c|}{67.1}          & 73.9             \\
\multicolumn{1}{l|}{Nemotron-4-340B-Reward}                                     & 92.0                 & 69.5              & \multicolumn{1}{c|}{69.9}          & 77.1             \\
\multicolumn{1}{l|}{Skywork-Reward-Llama-3.1-8B}                                & 92.5                 & 70.1              & \multicolumn{1}{c|}{69.3}          & 77.5             \\
\multicolumn{1}{l|}{infly/INF-ORM-Llama3.1-70B}                                 & \textbf{95.1}        & 70.9              & \multicolumn{1}{c|}{70.5}          & 78.8             \\ \midrule
\multicolumn{5}{l}{\textit{\textbf{Generative Reward Models}}}                                                                                                                     \\ \midrule
\multicolumn{1}{l|}{\cellcolor[HTML]{EFEFEF}JudgeLRM}                           & 75.2                 & 64.7              & \multicolumn{1}{c|}{53.1}          & 64.3             \\
\multicolumn{1}{l|}{Claude-3.5-sonnet-20240620}                                 & 84.2                 & 61.0              & \multicolumn{1}{c|}{70.6}          & 71.9             \\
\multicolumn{1}{l|}{Llama3.1-70B-Instruct}                                      & 84.0                 & 65.5              & \multicolumn{1}{c|}{68.9}          & 72.8             \\
\multicolumn{1}{l|}{Gemini-1.5-pro}                                             & 88.2                 & 75.2              & \multicolumn{1}{c|}{56.5}          & 73.3             \\
\multicolumn{1}{l|}{\cellcolor[HTML]{EFEFEF}DeepSeek-GRM-27B}                   & 86.0                 & --                & \multicolumn{1}{c|}{69.0}          & --               \\
\multicolumn{1}{l|}{\cellcolor[HTML]{EFEFEF}Self-taught-evaluator-llama3.1-70B} & 90.2                 & 71.4              & \multicolumn{1}{c|}{67.0}          & 76.2             \\
\multicolumn{1}{l|}{Skywork-Critic-Llama-3.1-70B}                               & 93.3                 & 71.9              & \multicolumn{1}{c|}{65.5}          & 76.9             \\
\multicolumn{1}{l|}{GPT-4o-0806}                                                & 86.7                 & 72.5              & \multicolumn{1}{c|}{\textbf{73.8}} & 77.7             \\
\multicolumn{1}{l|}{\cellcolor[HTML]{EFEFEF}RM-R1-Qwen-Instruct-32B}                                                & 91.4                 & 79.1              & \multicolumn{1}{c|}{\underline{73.0}} & 81.2             \\
\multicolumn{1}{l|}{\cellcolor[HTML]{EFEFEF}RM-R1-DeepSeek-Distilled-Qwen-32B}  & 90.9                 & { \underline{83.9} }        & \multicolumn{1}{c|}{69.8}          & {\underline{81.5}}       \\ \midrule
\multicolumn{1}{l|}{\cellcolor[HTML]{EFEFEF}\textbf{ReasonGRM-QwQ-32B (Ours)}}  & 92.3                 & \textbf{86.3}     & \multicolumn{1}{c|}{{ 71.3}}   & \textbf{83.3}   \\ \bottomrule
\end{tabular}}
\caption{\textbf{Performance comparison of ReasonGRM with various reward models on RewardBench, RM-Bench, and RMB benchmarks.} Scores in \textbf{bold} indicate the best performance, while \underline{underlined} scores indicate the second best. Models with a \hl{gray background} are specifically optimized for reasoning capabilities.}
\label{tab:main}
\end{table*}

\subsection{Reinforcement Learning with GRPO}
While Supervised Fine-Tuning in Stage 2 effectively teaches the model to replicate high-quality reasoning patterns, its discriminative boundaries may still be suboptimal for ambiguous or particularly challenging cases. To further sharpen the model's decision-making capabilities and maximize its performance on these difficult instances, we introduce a final reinforcement learning stage focused exclusively on hard cases.

The process involves two steps:

\textbf{1) Hard-Case Data Identification:} First, for each query $Q_j$ in the initial dataset $\mathcal{D}$, we use the $M_{\text{SFT}}$ model to infer $N$ answers. Let this set of $N$ answers for a specific $Q_j$ be denoted as $A_{\text{pred}} = \{\text{ans}_1, \text{ans}_2, \ldots, \text{ans}_N\}$. If $A_{\text{pred}}$ is not entirely correct, then this query $Q_j$ is included in the hard dataset $\mathcal{D}_{\text{hard}} = \{(R^{(k)}, A^{(k)})\}_{k=1}^{K}$. $A^{(k)}$ refers to the set of $N$ predicted answers $A_{\text{pred}}$ for the $k$-th hard query, and $R^{(k)}$ is its associated reference or rationale.

\textbf{2) Final RL Refinement:} Subsequently, we employ the GRPO to perform a final round of fine-tuning on the $M_{\text{SFT}}$. Crucially, this training phase uses only the hard-case dataset $\mathcal{D}_{\text{hard}}$. The reward mechanism remains consistent with Stage 1, providing an outcome-based reward based on the correctness of the final answer. By concentrating the reinforcement learning on the specific examples where the model is most likely to error, we can efficiently refine its judgment on the most critical decision boundaries. This final optimization step yields our fully-tuned model, ReasonGRM $M_{\text{grpo}}$.

\section{Experiments}
In this section, we conduct a series of experiments to comprehensively evaluate the effectiveness of our proposed $R\star$ metric and training pipeline and to demonstrate the superiority of the ReasonGRM. We will first introduce the experimental setup, including the datasets, benchmarks, and baseline models. Subsequently, we will present the main experimental results and conduct a comparative analysis of performance and methodology against current state-of-the-art models. Finally, we will dissect the contribution of each key component within ReasonGRM through detailed ablation studies.

\subsection{Experimental Setup}
\subsubsection{Dataset}
We use the \textbf{Skywork Reward Preference 80K v0.2}~\cite{liu2024skywork} dataset as our initial data source. Developed by Skywork AI, this high-quality preference alignment dataset provides approximately 80,000 rigorously selected, cross-domain samples. These samples were compiled by extracting data from publicly available web sources using specific strategies. The dataset encompasses a range of complex tasks, including mathematics, programming code, and logical reasoning.

\subsection{Benchmarks}
\begin{itemize}[label=\textbullet, wide, labelindent=0pt, leftmargin=*]
    \item \textbf{RewardBench}~\cite{lambert2024rewardbenchevaluatingrewardmodels}: This benchmark provides a foundational dataset for evaluating RMs. It consists of triplets of prompts, winning responses, and losing responses, covering diverse domains such as chat, reasoning, and safety. Reward-Bench is designed to assess RMs on challenging, structured, and out-of-distribution queries.
    \item \textbf{RM-Bench}~\cite{liu2024rm}: RM-Bench introduces an evaluation framework tailored for reward models. Its primary focus is on assessing a model's proficiency in discerning subtle content distinctions and its robustness to stylistic variations. This framework incorporates a `Style-Substance Evaluation Matrix' and a variety of accuracy metrics (e.g., simple, medium, and hard accuracy levels) to enable detailed, fine-grained evaluation.
    \item  \textbf{RMB}~\cite{zhou2024rmb}: RMB introduces a more comprehensive benchmark designed to evaluate the alignment capabilities of Reward Models. It encompassing over 49 distinct, fine-grained real-world scenarios. This benchmark integrates traditional pairwise comparisons with an innovative Best-of-N evaluation methodology. This design is intended to more effectively capture the practical efficacy of reward models in guiding alignment optimization.
\end{itemize}

\subsection{Implementation Details}
All models in this study were trained on a cluster of 4 nodes, each equipped with 8×NVIDIA A800-80G GPUs. The learning rate for all training tasks was uniformly set to $1 \times 10^{-6}$, with a global batch size of 256, a maximum prompt length of 8,192 tokens, and a response length limit of 57,344 tokens.

\begin{table}[]
\resizebox{\columnwidth}{!}{%
\begin{tabular}{@{}l|cccc|c@{}}
\toprule
\multicolumn{1}{c|}{\textbf{Method}} & \textbf{Chat}  & \textbf{Chat Hard} & \textbf{Safety} & \textbf{Reasoning} & \textbf{Score} \\ \midrule
QwQ-32B (Base)                   & 95.25          & 80.48              & 88.51           & 97.49              & 90.43          \\ \midrule
QwQ + \textbf{Zero-RL}                          & 91.62          & \textbf{86.51}     & 91.42           & 98.38              & 91.98          \\
QwQ + \textbf{R* SFT}                           & 92.60          & 86.18              & \textbf{91.82}  & 98.36              & 92.11          \\
{\ \ \ \ \ \ \ \ \ \ }+ \textbf{GRPO}                                & \textbf{96.09} & 83.55              & 90.81           & \textbf{98.74}     & \textbf{92.30} \\ \bottomrule
\end{tabular}%
}
\caption{Ablation study results for different training stages of ReasonGRM. Scores in \textbf{bold} indicate the highest performance.}
\label{tab:abl1}
\end{table}

\begin{table}[]
\resizebox{\columnwidth}{!}{%
\begin{tabular}{@{}r|cccc|c@{}}
\toprule
\multicolumn{1}{c|}{\textbf{Model}} & \textbf{Chat}  & \textbf{Chat Hard} & \textbf{Safety} & \textbf{Reasoning} & \textbf{Score} \\ \midrule
\multicolumn{1}{l|}{Llama3.1-8B}    & 91.83          & 51.97              & 78.92           & 62.48              & 71.30          \\
w/ Random-SFT                       & 93.78          & 73.79              & 87.40           & 72.42              & 81.85          \\
\textbf{w/ R*-SFT}                  & \textbf{93.58} & \textbf{74.01}     & \textbf{86.89}  & \textbf{73.95}     & \textbf{82.11} \\ \midrule
\multicolumn{1}{l|}{Qwen2.5-7B}     & \textbf{96.65} & 60.09              & 81.69           & 80.67              & 79.78          \\
w/ Random-SFT                       & 92.95          & 74.95              & 84.70           & 83.88              & 84.12          \\
\textbf{w/ R*-SFT}                  & 93.51          & \textbf{76.48}     & \textbf{85.30}  & \textbf{84.38}     & \textbf{84.92} \\ \midrule
\multicolumn{1}{l|}{Qwen2.5-14B}    & \textbf{95.39} & 67.32              & 84.73           & 81.66              & 82.28          \\
w/ Random-SFT                       & 94.62          & 80.65              & \textbf{87.84}  & 92.12              & 88.81          \\
\textbf{w/ R*-SFT}                  & 95.25          & \textbf{81.69}     & 87.77           & \textbf{93.78}     & \textbf{89.62} \\ \bottomrule
\end{tabular}%
}
\caption{\textbf{Effectiveness of the $R^\star$ metric in different models.} This table compares the performance of three models on RewardBench. For each model, we present its performance after SFT with randomly sampled data versus SFT with data filtered by $R^\star$ metric. Scores in \textbf{bold} indicate the highest performance of each model.}
\label{tab:abl2}
\end{table}

\subsection{Main Results}
\subsubsection{Overall performance}
Our proposed ReasonGRM demonstrates exceptional performance across three challenging benchmarks for reward models, as detailed in Table \ref{tab:main}. ReasonGRM achieves a SOTA average performance of 83.3, surpassing all competing SRMs and other GRMs evaluated. This superior result is particularly noteworthy as ReasonGRM was developed without reliance on proprietary large-scale models for distillation or guidance. The consistent high performance across RewardBench, RM-Bench, and RMB robustly validates our hypothesis: enhancing the intrinsic reasoning capabilities of a reward model is a highly effective pathway to significantly improving its preference modeling accuracy.

\subsubsection{Importance of general reasoning ability.}
As shown in Table \ref{tab:main}, some SRMs (e.g., infly/INF-ORM-Llama3.1-70B), despite achieving an excellent score as high as 95.1 on RewardBench (2.8 points higher than ReasonGRM), experienced a steep performance decline to 70.9 points on RM-Bench, which places greater emphasis on discerning subtle content differences and style robustness. This score was, conversely, 15.4 points lower than that of ReasonGRM. This phenomenon of significant performance degradation across different benchmarks is prevalent among most SRMs and GRMs lacking in-depth reasoning optimization, with the core reason being RM-Bench's high demands on the model's intrinsic reasoning capabilities.

In contrast, our ReasonGRM (based on the reasoning-specialized QwQ model) and RM-R1-DeepSeek-Distilled-Qwen-32B (based on DeepSeek-Distilled-Qwen-32B) exhibited stronger robustness and a smaller performance decline on RM-Bench; notably, ReasonGRM achieved the highest score on this benchmark. This comparison starkly reveals that \textbf{strong general reasoning ability is the cornerstone for reward models to maintain high-level performance when addressing complex and nuanced evaluation tasks}. Therefore, further specializing LRMs endowed with strong general reasoning capabilities into dedicated reward models like ReasonGRM holds significant value for advancing the field.\\

\subsection{Ablation Study}
\subsubsection{Contributions of each stage}
Starting from the QwQ base model, we incrementally applied the distinct training stages of ReasonGRM, evaluating each on the Reward-Bench benchmark. This approach enabled a quantitative evaluation of the performance contributions from: (1) the initial Zero-RL adaptation, (2) SFT with $R^\star$ Refinement, and (3) the final GRPO reinforcement learning refinement. The specific evolution of performance is detailed in Table \ref{tab:abl1}.

\begin{itemize}[label=\textbullet, wide, labelindent=0pt, leftmargin=*]
    \item \textbf{Zero-RL substantially boosts initial performance}: The initial GRPO training of the base QwQ model, conducted on data devoid of explicit reasoning processes, yielded ReasonGRM-QwQ-Zero. This model demonstrated a 1.55\% uplift in accuracy on Reward-Bench compared to the original QwQ. This outcome suggests that \textbf{Zero-RL effectively enables an initial calibration of the model to the fundamental paradigms of the preference task}, thereby establishing a robust foundation for its subsequent comprehension and generation of high-quality reasoning paths.

    \item \textbf{$R^\star$-Guided SFT leverages high-quality reasoning paths to further elevate model capabilities.}: Subsequently, applying SFT to QwQ with high-quality reasoning paths—generated by ReasonGRM-QwQ-Zero and refined via the $R^\star$ metric—produced ReasonGRM-QwQ-sft, which achieved a further performance increase of 1.68\%. This result \textbf{underscores the critical role of high-quality, logically coherent reasoning exemplars in guiding the model to learn effective discriminative strategies}.

    \item \textbf{GRPO refinement on challenging samples culminates in the highest achieved performance.}: Finally, optimizing ReasonGRM-QwQ-sft via GRPO on challenging samples culminated in the ReasonGRM-QwQ-grpo model, registering further accuracy improvements. This demonstrates that after the model has developed substantial reasoning and discriminative capabilities, \textbf{fine-tuning through reinforcement learning on difficult samples serves to effectively sharpen its decision boundaries and maximize its comprehensive performance on target tasks.}
\end{itemize}

\subsubsection{Effectiveness of $R^\star$}
To perform a more direct and comprehensive evaluation of the intrinsic efficacy of the $R^\star$ metric and its universal applicability across different models, we designed a comparative experiment against a random sampling strategy. In this experimental setup, we consistently utilized data previously generated by the QwQ-Zero model. Then, after ensuring that all chosen reasoning paths yielded correct final answers, we employed two distinct strategies for filtering the Supervised Fine-Tuning (SFT) training data: $R^\star$ sampling and random sampling. The data filtered by each strategy was then used to instruction fine-tune three distinct base models, varying in architecture and/or scale: Llama3.1-8B, Qwen2.5-7B, and Qwen2.5-14B. All models fine-tuned with these datasets were then evaluated on the RewardBench benchmark, with detailed results presented in Table \ref{tab:abl2}.

The results consistently indicated that, for all base models tested, models trained on data selected via the $R^\star$ metric significantly outperformed those trained on data chosen by random sampling. This strongly validates the effectiveness of the $R^\star$ metric and its generalizable value across different models.

\section{Conclusion}
This paper addresses the challenge of deficient reasoning capabilities in GRMs by introducing ReasonGRM alongside a systematic solution. We leveraged the $R^{*}$ metric to facilitate the quantitative evaluation and selection of generated reasoning data. Furthermore, we designed a comprehensive, multi-stage workflow that encompasses data generation, filtering, and training. Comparisons of ReasonGRM against other reward models on mainstream benchmarks demonstrate its SOTA performance, thereby validating the efficacy of our training methodology. Ablation studies further substantiated the broad applicability of the $R^{*}$ metric and underscored the critical contributions of both the quality of reasoning processes and each distinct training stage to the model's ultimate performance. Future research avenues include exploring the application of the $R^{*}$ metric to a broader spectrum of reasoning tasks and investigating more sophisticated methodologies for generating and selecting high-quality reasoning paths.

\section*{Limitations}
\subsection*{Evaluation of Reasoning Beyond Benchmarks}
Although ReasonGRM has been validated on several mainstream benchmarks, assessing its reasoning capabilities in more open-ended, real-world scenarios or across a broader spectrum of complex, multi-hop reasoning tasks remains an important area for continued investigation.

\subsection*{Scope of Application of $R^\star$}
While $R^\star$ demonstrates considerable efficacy on datasets featuring clearly defined question-answer (QA) pairs, its applicability to datasets with open-ended answers remains challenging. This presents a limitation to its widespread adoption. Future work could investigate adapting the $R^\star$ metric for effective application to datasets characterized by open-ended answers.




\bibliography{anthology,custom}
\bibliographystyle{acl_natbib}

\appendix
\newpage
\section{Prompt Template}
\label{sec:appendix1}
\begin{figure*}[htbp]
  \centering 
  \includegraphics[width=0.85\linewidth]{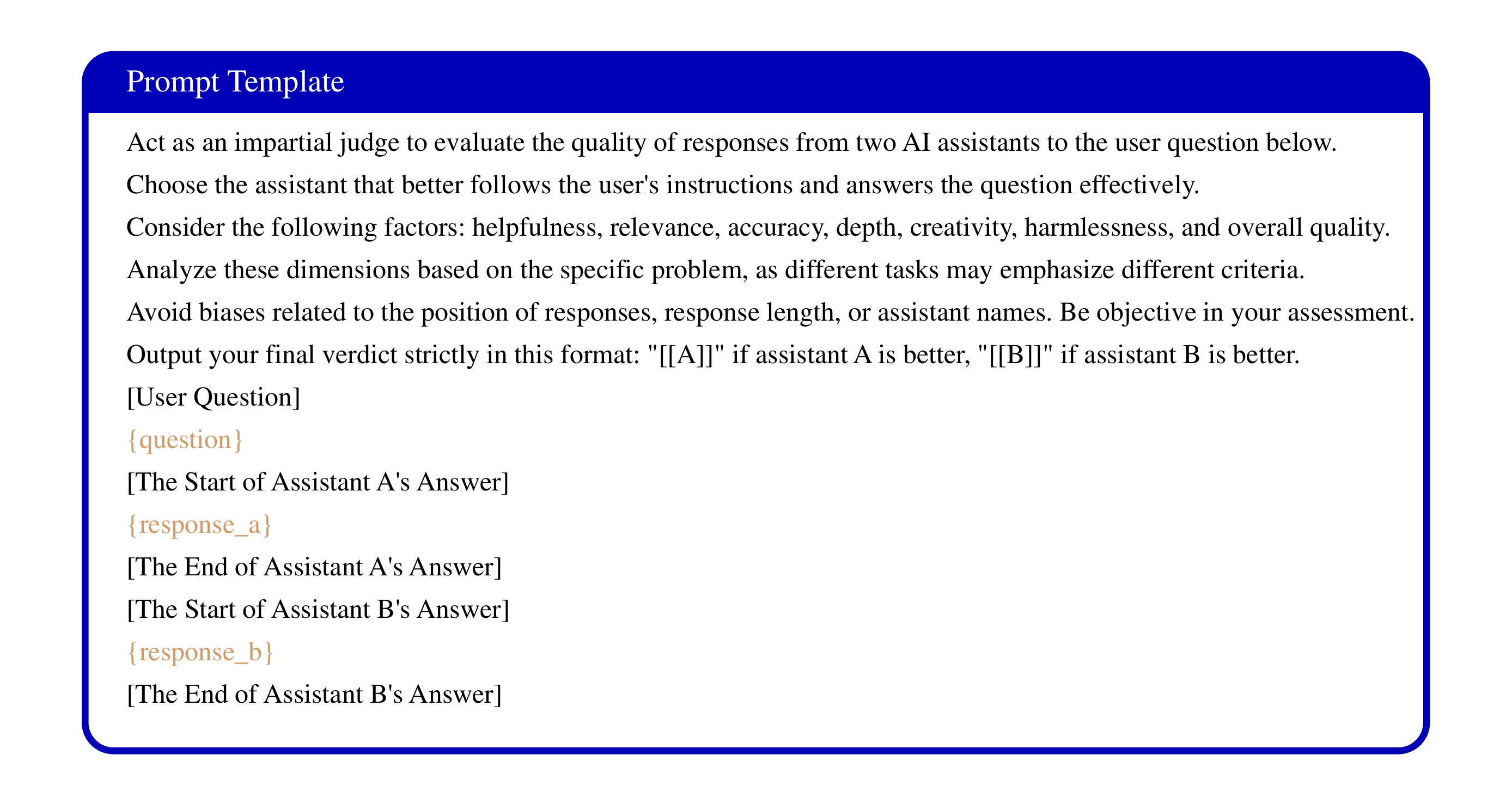}
  \vspace{-10pt}
  \caption{The prompt template for Experiments}
  \label{fig:prompt} 
\end{figure*}

The prompt template we used for data generation, training, and evaluation is shown in Figure ~\ref{fig:prompt}.

\section{Baselines}
\label{sec:appendix2}
We compare ReasonGRM with a array of state-of-the-art reward models. These baselines are broadly categorized into two groups: Scalar Reward Models (SRMs) and Generative Reward Models (GRMs).
\subsection{Scalar Reward Models}
SRMs are trained to output a single scalar value representing the quality of a response. They treat reward modeling as a regression or classification task but typically lack explicit reasoning capabilities. This category includes:
\begin{itemize}
    \item Skywork-Reward-Gemma-2-27B~\citep{liu2024skywork}
    \item Skywork-Reward-Llama-3.1-8B~\citep{liu2024skywork}
    \item Internlm2-7b-reward~\citep{cai2024internlm2}
    \item Nemotron-4-340B-Reward~\citep{adler2024nemotron}
    \item INF-ORM-Llama3.1-70B~\cite{INF-ORM-Llama3.1-70B}
\end{itemize}

\subsection{Scalar Reward Models}
GRMs leverage the generative capabilities of Large Language Models to produce text-based judgments, which often include critiques or explanations for their decisions. This diverse category includes both powerful, general-purpose models used directly as evaluators, as well as models that have been specifically fine-tuned to enhance their reasoning abilities for evaluation tasks. The models we compare against are:
\begin{itemize}
    \item Claude-3.5-sonnet~\citep{anthropic2024claude}
    \item Gemini-1.5-pro~\citep{reid2024gemini}
    \item GPT-4o~\citep{achiam2023gpt4, hurst2024gpt}
    \item Llama3.1-70B-Instruct~\citep{dubey2024llama}
    \item Skywork-Critic-Llama-3.1-70B~\cite{skyworkcritic2024}
    \item JudgeLRM~\citep{chen2025judgelrm}
    \item DeepSeek-GRM-27B~\citep{liu2025inferencetimescalinggeneralistreward}
    \item Self-taught-evaluator-llama3.1-70B~\citep{wang2024self}
    \item RM-R1~\cite{chen2025rmr1rewardmodelingreasoning}
\end{itemize}

\section{Case Study}
\label{sec:appendix3}
To qualitatively illustrate ReasonGRM's superior reasoning capabilities, we present a detailed case study. This case illustrated the reasoning process of the QwQ with ReasonGRM when evaluating responses with subtle factual inaccuracies. The user prompt and the two candidate responses (A and B) that serve as the input for the reward models are shown in Figure 1. Response A was designated as the Chosen response, and Response B as the Rejected one. The reasoning process in Figure 3 illustrates the QwQ's struggle with conflicting information. The model vacillates multiple times (e.g., "Wait, but...", "Let me check again."). Although it correctly identifies the internal contradiction in Response B's education timeline ("B mentions 2-3 years for education, but then notes that 150 hours is about five years"), it ultimately gets distracted by superficial features like "comprehensiveness" and "detail". By overlooking this critical flaw in accuracy, the model reaches an incorrect final judgment, favoring Response B. In contrast, the reasoning process of ReasonGRM is clear and decisive. The model immediately identifies the "discrepancy" and "inconsistency" in Response B's education timeline and correctly weighs this as the most critical factor for its judgment. The entire reasoning process is logically coherent and avoids vacillation, leading to the correct choice based on the core principles of "accuracy" and "clarity." This demonstrates ReasonGRM's superior ability to discern substantive content quality over superficial presentation. 

\section{Full Experimental Results}
\label{sec:appendix4}
This section provides the full, detailed results of our experiments on the RewardBench, RM-Bench, and RMB benchmarks.

\begin{figure*}[htbp]
  \centering 
  \includegraphics[width=0.9\linewidth]{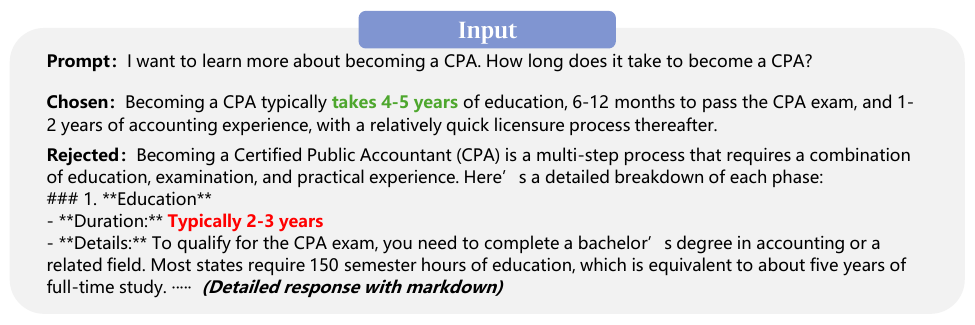}
  \caption{Case study input: The user asks about the time required to become a CPA. Response A (Chosen) gives a direct summary, while Response B (Rejected) provides a detailed breakdown but contains a factual contradiction regarding the education duration.}
  \label{fig:case_input} 
\end{figure*}

\begin{figure*}[htbp]
  \centering 
  \includegraphics[width=0.9\linewidth]{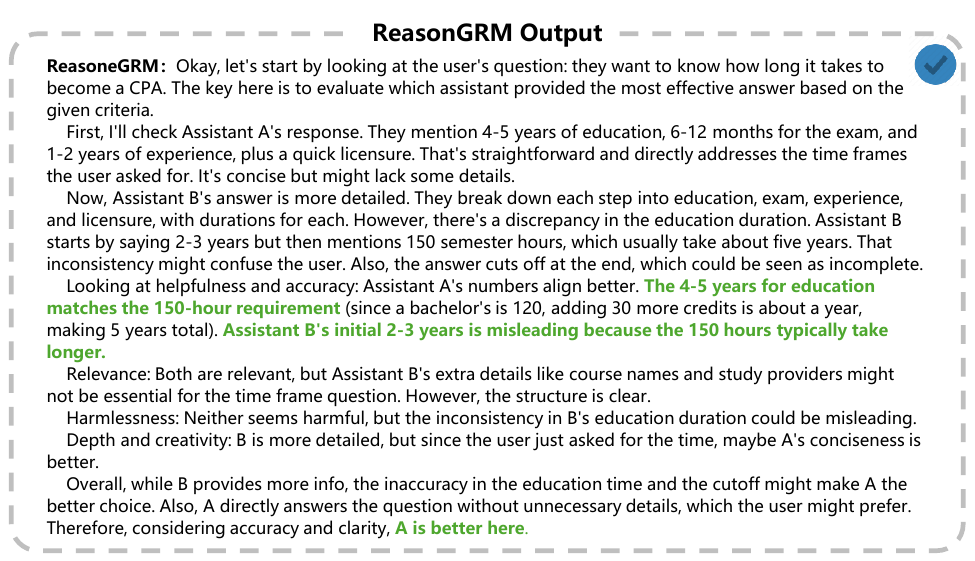}
  \caption{Reasoning process of ReasonGRM for the case study. The model quickly and accurately identifies the inconsistency in Response B's timeline ("discrepancy") and correctly concludes that Response A is better due to superior accuracy and clarity.}
  \label{fig:case_grm} 
\end{figure*}

\begin{figure*}[htbp]
  \centering 
  \includegraphics[width=1.0\linewidth]{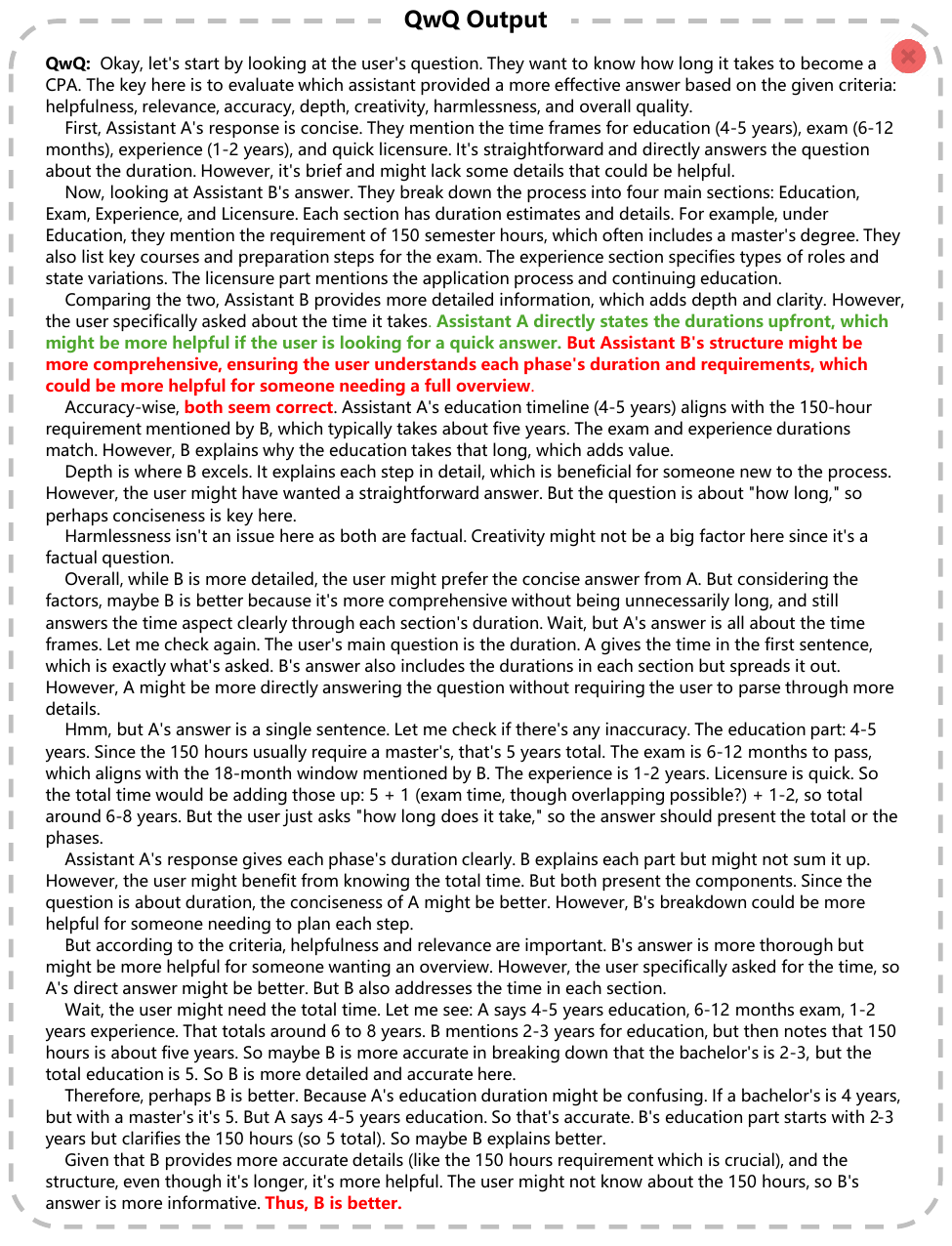}
  \caption{Reasoning process of the baseline QwQ for the case study. The model shows significant vacillation. Although it notices the contradiction in Response B, it ultimately gets distracted by superficial details ("more comprehensive") and makes the wrong choice.}
  \label{fig:case_qwq} 
\end{figure*}


\begin{table*}[htbp]
\centering
\begin{tabular}{@{}lccccc@{}}
\toprule
\multicolumn{1}{l|}{\textbf{Models}}                                           & \textbf{Chat} & \textbf{Chat Hard} & \textbf{Safety} & \multicolumn{1}{c|}{\textbf{Reasoning}} & \textbf{Score} \\ \midrule
\multicolumn{5}{l}{\textit{\textbf{Scalar Reward Models}}}                                                                                                                      &                \\ \midrule
\multicolumn{1}{l|}{Skywork-Reward-Gemma-2-27B}                                & 95.8          & \textbf{91.4}      & 92.0            & \multicolumn{1}{c|}{96.1}               & \ul {93.8}     \\
\multicolumn{1}{l|}{Internlm2-7b-reward}                                       & \textbf{98.6} & 66.7               & 88.3            & \multicolumn{1}{c|}{92.8}               & 86.6           \\
\multicolumn{1}{l|}{Nemotron-4-340B-Reward}                                    & 95.8          & 87.1               & 92.2            & \multicolumn{1}{c|}{93.6}               & 92.2           \\
\multicolumn{1}{l|}{Skywork-Reward-Llama-3.1-8B}                               & 95.8          & 87.3               & 90.6            & \multicolumn{1}{c|}{96.2}               & 92.5           \\
\multicolumn{1}{l|}{INF-ORM-Llama3.1-70B}                                      & 96.6          & \ul {91.0}         & \textbf{93.6}   & \multicolumn{1}{c|}{\textbf{99.1}}      & \textbf{95.1}  \\ \midrule
\multicolumn{5}{l}{\textit{\textbf{Generative Reward Models}}}                                                                                                                  &                \\ \midrule
\multicolumn{1}{l|}{\cellcolor[HTML]{EFEFEF}JudgeLRM}                          & 92.9          & 56.4               & 78.2            & \multicolumn{1}{c|}{73.6}               & 75.2           \\
\multicolumn{1}{l|}{Claude-3.5-sonnet-20240620}                                & 96.4          & 74.0               & 81.6            & \multicolumn{1}{c|}{84.7}               & 84.2           \\
\multicolumn{1}{l|}{Llama3.1-70B-Instruct}                                     & \ul {97.2}    & 70.2               & 82.8            & \multicolumn{1}{c|}{86.0}               & 84.0           \\
\multicolumn{1}{l|}{Gemini-1.5-pro}                                            & 92.3          & 80.6               & 87.9            & \multicolumn{1}{c|}{92.0}               & 88.2           \\
\multicolumn{1}{l|}{\cellcolor[HTML]{EFEFEF}DeepSeek-GRM-27B}                  & 94.1          & 78.3               & 88.0            & \multicolumn{1}{c|}{83.8}               & 86.0           \\
\multicolumn{1}{l|}{Self-taught-evaluator-llama3.1-70B}                        & 96.9          & 85.1               & 89.6            & \multicolumn{1}{c|}{88.4}               & 90.0           \\
\multicolumn{1}{l|}{Skywork-Critic-Llama-3.1-70B}                              & 96.6          & 87.9               & \ul {93.1}      & \multicolumn{1}{c|}{95.5}               & 93.3           \\
\multicolumn{1}{l|}{GPT-4o-0806}                                               & 96.1          & 76.1               & 86.6            & \multicolumn{1}{c|}{88.1}               & 86.7           \\
\multicolumn{1}{l|}{\cellcolor[HTML]{EFEFEF}RM-R1-Qwen-Instruct-32B}           & 95.3          & 83.1               & 91.9            & \multicolumn{1}{c|}{95.2}               & 91.4           \\
\multicolumn{1}{l|}{\cellcolor[HTML]{EFEFEF}RM-R1-DeepSeek-Distilled-Qwen-32B} & 95.3          & 80.3               & 91.1            & \multicolumn{1}{c|}{96.8}               & 90.9           \\ \midrule
\multicolumn{1}{l|}{\cellcolor[HTML]{EFEFEF}ReasonGRM-QwQ-32B (Ours)}          & 96.9          & 83.6              & 90.8           & \multicolumn{1}{c|}{\ul {98.7}}        & 92.3           \\ \bottomrule
\end{tabular}
\caption{Full experimental results on the RewardBench benchmark. Scores in bold indicate the best performance in a category, while underlined scores are the second best. Models with a gray background are specifically optimized for reasoning capabilities.}
\label{tab:rewardbench_full}
\end{table*}

\begin{table*}[htbp]
\centering
\resizebox{2\columnwidth}{!}{%
\begin{tabular}{@{}lcccccccc@{}}
\toprule
\multicolumn{1}{l|}{\textbf{Models}}                                           & \textbf{Chat} & \textbf{Math} & \textbf{Code} & \multicolumn{1}{c|}{\textbf{Safety}} & \textbf{Easy} & \multicolumn{1}{l}{\textbf{Normal}} & \multicolumn{1}{l|}{\textbf{Hard}} & \multicolumn{1}{l}{\textbf{Avg}} \\ \midrule
\multicolumn{5}{l}{\textit{\textbf{Scalar Reward Models}}}                                                                                                            &               & \multicolumn{1}{l}{}                & \multicolumn{1}{l}{}               & \multicolumn{1}{l}{}             \\ \midrule
\multicolumn{1}{l|}{Skywork-Reward-Gemma-2-27B}                                & 69.5          & 54.7          & 53.2          & \multicolumn{1}{c|}{91.9}            & 78.0          & 69.2                                & \multicolumn{1}{c|}{54..9}         & 67.3                             \\
\multicolumn{1}{l|}{Internlm2-7b-reward}                                       & 61.7          & 71.4          & 49.7          & \multicolumn{1}{c|}{85.5}            & 85.4          & 70.7                                & \multicolumn{1}{c|}{45.1}          & 67.1                             \\
\multicolumn{1}{l|}{Nemotron-4-340B-Reward}                                    & 71.2          & 59.8          & 59.4          & \multicolumn{1}{c|}{87.5}            & 81.0          & 71.4                                & \multicolumn{1}{c|}{56.1}          & 69.5                             \\
\multicolumn{1}{l|}{Skywork-Reward-Llama-3.1-8B}                               & 69.5          & 60.6          & 54.5          & \multicolumn{1}{c|}{\textbf{95.7}}   & 89.0          & 74.7                                & \multicolumn{1}{c|}{46.6}          & 70.1                             \\
\multicolumn{1}{l|}{INF-ORM-Llama3.1-70B}                                      & 66.3          & 65.6          & 56.8          & \multicolumn{1}{c|}{94.8}            & \textbf{91.8} & 76.1                                & \multicolumn{1}{c|}{44.8}          & 70.9                             \\ \midrule
\multicolumn{5}{l}{\textit{\textbf{Generative Reward Models}}}                                                                                                        &               &                                     &                                    &                                  \\ \midrule
\multicolumn{1}{l|}{\cellcolor[HTML]{EFEFEF}JudgeLRM}                          & 59.9          & 59.9          & 51.9          & \multicolumn{1}{c|}{87.3}            & 73.2          & 76.2                                & \multicolumn{1}{c|}{54.8}          & 64.7                             \\
\multicolumn{1}{l|}{Claude-3.5-sonnet-20240620}                                & 62.5          & 62.6          & 54.4          & \multicolumn{1}{c|}{64.4}            & 73.8          & 63.4                                & \multicolumn{1}{c|}{45.9}          & 61.0                             \\
\multicolumn{1}{l|}{Llama3.1-70B-Instruct}                                     & 64.3          & 67.3          & 47.5          & \multicolumn{1}{c|}{83.0}            & 74.7          & 67.8                                & \multicolumn{1}{c|}{54.1}          & 65.5                             \\
\multicolumn{1}{l|}{Gemini-1.5-pro}                                            & 71.6          & 73.9          & 63.7          & \multicolumn{1}{c|}{91.3}            & 83.1          & 77.6                                & \multicolumn{1}{c|}{64.7}          & 75.2                             \\
\multicolumn{1}{l|}{\cellcolor[HTML]{EFEFEF}DeepSeek-GRM-27B}                  & -             & -             & -             & \multicolumn{1}{c|}{-}               & -             & -                                   & \multicolumn{1}{c|}{-}             & -                                \\
\multicolumn{1}{l|}{Self-taught-evaluator-llama3.1-70B}                        & 73.4          & 65.7          & 56.3          & \multicolumn{1}{c|}{90.4}            & 80.2          & 74.5                                & \multicolumn{1}{c|}{59.7}          & 71.5                             \\
\multicolumn{1}{l|}{Skywork-Critic-Llama-3.1-70B}                              & 71.4          & 64.6          & 56.8          & \multicolumn{1}{c|}{94.8}            & 85.6          & 73.7                                & \multicolumn{1}{c|}{56.5}          & 71.9                             \\
\multicolumn{1}{l|}{GPT-4o-0806}                                               & 67.2          & 67.5          & 63.6          & \multicolumn{1}{c|}{91.7}            & 83.4          & 75.6                                & \multicolumn{1}{c|}{58.7}          & 72.5                             \\
\multicolumn{1}{l|}{\cellcolor[HTML]{EFEFEF}RM-R1-Qwen-Instruct-32B}           & \ul {75.3}    & 80.2          & 66.8          & \multicolumn{1}{c|}{93.9}            & 86.3          & 80.5                                & \multicolumn{1}{c|}{70.4}          & 79.1                             \\
\multicolumn{1}{l|}{\cellcolor[HTML]{EFEFEF}RM-R1-DeepSeek-Distilled-Qwen-32B} & 74.2          & \ul {91.8}    & \ul {74.1}    & \multicolumn{1}{c|}{\ul {95.4}}      & 89.5          & \ul {85.4}                          & \multicolumn{1}{c|}{\ul {76.7}}    & \ul {83.9}                       \\ \midrule
\multicolumn{1}{l|}{\cellcolor[HTML]{EFEFEF}ReasonGRM-QwQ-32B (Ours)}          & \textbf{82.1} & \textbf{94.7} & \textbf{80.3} & \multicolumn{1}{c|}{90.0}            & \ul {90.1}    & \textbf{88.5}                       & \multicolumn{1}{c|}{\textbf{81.8}} & \textbf{86.8}                    \\ \bottomrule
\end{tabular}}
\caption{Full experimental results on the RM-Bench benchmark. This benchmark assesses a model's ability to discern subtle content differences and its robustness to stylistic variations across various difficulty levels.}
\label{tab:rmbench_full}
\end{table*}


\begin{table*}[htbp]
\centering
\begin{tabular}{@{}lccccc@{}}
\toprule
\multirow{2}{*}[-5pt]{\textbf{\ \ \ Models}} & \multicolumn{2}{|c}{\textbf{Helpfulness}} & \multicolumn{2}{c|}{\textbf{Harmlessness}} & \\ 
\cmidrule(lr){2-3} \cmidrule(lr){4-5}
& \multicolumn{1}{|c}{\textbf{BoN}} & \textbf{Pairwise} & \textbf{BoN} & \multicolumn{1}{c|}{\textbf{Pairwise}} & \multirow{-2}{*}[2pt]{\textbf{Avg}} \\ 
\midrule
\multicolumn{5}{l}{\textit{\textbf{Scalar Reward Models}}}                                                                                                                         &                                \\ \midrule
\multicolumn{1}{l|}{Skywork-Reward-Gemma-2-27B}                                & 47.2              & 65.3                 & 56.1          & \multicolumn{1}{c|}{72.1}              & 60.2                           \\
\multicolumn{1}{l|}{Internlm2-7b-reward}                                       & 62.6              & 78.2                 & 56.3          & \multicolumn{1}{c|}{71.2}              & 67.1                           \\
\multicolumn{1}{l|}{Nemotron-4-340B-Reward}                                    & -                 & -                    & -             & \multicolumn{1}{c|}{-}                 & 69.9                           \\
\multicolumn{1}{l|}{Skywork-Reward-Llama-3.1-8B}                               & \ul {69.5}        & 60.6                 & 54.5          & \multicolumn{1}{c|}{\textbf{95.7}}     & 70.1                           \\
\multicolumn{1}{l|}{INF-ORM-Llama3.1-70B}                                      & 65.0              & 79.8                 & 60.7          & \multicolumn{1}{c|}{76.7}              & 70.5                           \\ \midrule
\multicolumn{5}{l}{\textit{\textbf{Generative Reward Models}}}                                                                                                                     &                                \\ \midrule
\multicolumn{1}{l|}{\cellcolor[HTML]{EFEFEF}JudgeLRM}                          & 36.3              & 69.9                 & 36.3          & \multicolumn{1}{c|}{67.4}              & 53.1                           \\
\multicolumn{1}{l|}{Claude-3.5-sonnet-20240620}                                & \textbf{70.5}     & \textbf{83.8}        & 51.8          & \multicolumn{1}{c|}{76.4}              & 70.6                           \\
\multicolumn{1}{l|}{Llama3.1-70B-Instruct}                                     & 64.8              & 81.1                 & 55.8          & \multicolumn{1}{c|}{73.9}              & 68.9                           \\
\multicolumn{1}{l|}{Gemini-1.5-pro}                                            & 53.6              & 76.3                 & 29.9          & \multicolumn{1}{c|}{66.1}              & 56.5                           \\
\multicolumn{1}{l|}{\cellcolor[HTML]{EFEFEF}DeepSeek-GRM-27B}                  & 62.3              & 80.5                 & 57.0          & \multicolumn{1}{c|}{76.1}              & 69.0                           \\
\multicolumn{1}{l|}{Self-taught-evaluator-llama3.1-70B}                        & 61.6              & 78.6                 & 54.6          & \multicolumn{1}{c|}{73.3}              & 67.0                           \\
\multicolumn{1}{l|}{Skywork-Critic-Llama-3.1-70B}                              & 64.0              & 75.3                 & 61.4          & \multicolumn{1}{c|}{61.4}              & 65.5                           \\
\multicolumn{1}{l|}{GPT-4o-0806}                                               & 63.9              & \ul {81.5}           & \textbf{68.2} & \multicolumn{1}{c|}{\ul {81.4}}        & \textbf{73.8}                           \\
\multicolumn{1}{l|}{\cellcolor[HTML]{EFEFEF}RM-R1-Qwen-Instruct-32B}           & 63.6              & 79.1                 & \textbf{68.2} & \multicolumn{1}{c|}{80.9}              & \ul{73.0}                  \\
\multicolumn{1}{l|}{\cellcolor[HTML]{EFEFEF}RM-R1-DeepSeek-Distilled-Qwen-32B} & 62.0              & 78.2                 & 61.8          & \multicolumn{1}{c|}{77.1}              & 69.8                           \\ \midrule
\multicolumn{1}{l|}{\cellcolor[HTML]{EFEFEF}ReasonGRM-QwQ-32B (Ours)}          & 62.3              & 77.2                 & 65.9          & \multicolumn{1}{c|}{80.0}              & {71.3}                     \\ \bottomrule
\end{tabular}
\caption{Full experimental results on the RMB benchmark. RMB evaluates models on fine-grained real-world scenarios using both Best-of-N (BoN) and Pairwise comparison methodologies for Helpfulness and Harmlessness.}
\label{tab:rmb_full}
\end{table*}

\end{document}